\documentclass[final]{cvpr}
\usepackage{times}
\usepackage{soul}
\usepackage{url}
\usepackage[hidelinks]{hyperref}
\usepackage[utf8]{inputenc}
\usepackage[small]{caption}
\usepackage{graphicx}
\usepackage{amsmath}
\usepackage{amsthm}
\usepackage{booktabs}
\usepackage{algorithm}
\usepackage{algorithmic}
\usepackage{subfig}
\usepackage{color}
\begin{document}
\title{Trear: Transformer-based RGB-D Egocentric Action Recognition}

\author{Xiangyu Li,
        Yonghong Hou,
        Pichao Wang,
        Zhimin Gao,
        Mingliang Xu,
        and Wanqing Li
\thanks{Corresponding authors: Zhimin Gao and Pichao Wang}
\thanks{X. Li and Y. Hou are with School of Electronic Information Engineering, Tianjing University, Tianjin, China (e-mail: lixiangyu\_1008@tju.edu.cn; houroy@tju.edu.cn).}
\thanks{P. Wang is with DAMO Academy, Alibaba Group (U.S.), Bellevue, USA (email: pichaowang@gmail.com).}
\thanks{Z. Gao and M. Xu are with School of Information Engineering, Zhengzhou University, Zhengzhou, China (e-mail: iegaozhimin@zzu.edu.cn; iexumingliang@zzu.edu.cn).}
\thanks{W. Li is with Advanced Multimedia Research Lab, University of Wollongong, Wollongong, Australia (email: wanqing@uow.edu.au).}}

\maketitle

\begin{abstract}
In this paper, we propose a  \textbf{Tr}ansformer-based RGB-D \textbf{e}gocentric \textbf{a}ction \textbf{r}ecognition framework, called Trear. It consists of two modules, inter-frame attention encoder and mutual-attentional fusion block. Instead of using optical flow or recurrent units, we adopt self-attention mechanism to model the temporal structure of the data from different modalities. Input frames are cropped randomly to mitigate the effect of the data redundancy. Features from each modality are interacted through the proposed fusion block and combined through a simple yet effective fusion operation to produce a joint RGB-D representation. Empirical experiments on two large egocentric RGB-D datasets, THU-READ and FPHA, and one small dataset, WCVS, have shown that the proposed method outperforms the state-of-the-art results by a large margin.\end{abstract}

\section{Introduction}
\label{sec:intro}
With the popularity of the wearable equipment (e.g. GoPro and VR helmet), recognition of human activities from egocentric videos has attracted much attention due to its wide research and practical applications, such as Robotics, VR/AR, etc. Recently, deep learning is widely applied to many computer vision tasks with promising results which promotes researchers to employ Convolutional Neural Networks (CNNs) or Recurrent Neural Networks (RNNs) in egocentric/third-person view action recognition \cite{lsta,attention,ma2016going,singh, yan2018multibranch, lee2018recognition}.  While promising, most of these methods are based on the single RGB modality, do not take the combination of multiple heterogeneous modalities, e.g. RGB and depth, into consideration. However, each modality has its own characteristic. Depth modality carries rich 3D structure information, shows insensitive to the illumination changes, and it lacks the vital texture appearance information, while RGB modality is vice versa. For conventional third person action recognition, RGB-D based methods \cite{wangAAAI,wangcv,kong2015bilinear} have been widely proposed to exploit the complementary characteristics of both modalities, while for egocentric action recognition, there are still few studies~\cite{THU-READ}. This paper focuses on RGB-D based egocentric action recognition and explores a novel framework to learn a conjoint representation of both modalities.

\par Compared to the third person action recognition, egocentric action is more fine-grained and it requires to classify both the motion performed by the subject and objects being manipulated (e.g. close juice bottle, close milk, pour milk, pour wine, etc.). Thus, it's essential to encode the spatial-temporal relation information of the action clips. Previous works either utilize the optical flow to exploit motion information via a two-stream network \cite{two-stream} or adopt a Convolutional Long Short-Term Memory (ConvLSTM) network for spatio-temporal encoding \cite{lsta,attention}. However, they either can only model short-term motion or only consider temporal structure sequentially as the activity progresses. Based on this observation, the recently proposed Transformer \cite{transformer} inspires us to employ it in RGB-D egocentric action recognition due to its strong capability of sequence modeling in NLP (e.g. language translation) tasks, parallelness in processing the input, and ability in building long-range dependencies through self-attention mechanism.

\par This paper proposes a novel transformer-based egocentric action recognition framework. It consists of two modules, inter-frame attention encoder and mutual-attentional fusion block. Data from each modality is first encoded through an attention encoder to build an intra-modality temporal structure, and then features are incorporated through the fusion block to produce a cross-modal representation. For the inter-frame attention encoder, we adopt a standard transformer encoder consisting of self-attention and feed-forward layers. By mimicking the language translation task, each sampled image (or depth map) in an action video is treated as a word and dependencies with other words is constructed using a self-attention mechanism. Due to the context redundancy among the sampled images, it's inefficient to conduct attention calculation. Thus, we propose to crop regions randomly from each image, where the different regions are interacted through the encoder to enhance the spatial correlation. Further, a mutual-attentional fusion block is proposed to learn joint representation for classification. In this block, self-attention layer is extended to a mutual-attention layer, where features from different modalities interact. Features after going through mutual-attention layer are fused via a simple operation to produce the cross-modal representation for classification. Our method is extensively evaluated on two large RGB-D egocentric datasets, THU-READ \cite{THU-READ} and FPHA \cite{garcia2018first}, and a small WCVS \cite{WCVS} dataset. Experiments results show that the proposed method achieves the state-of-the-art results and outperforms the existing methods by a large margin.

\par Our contributions can be summarized as follows:
\begin{itemize}
\item Transformer encoder is adopted to model the temporal contextual information over the action period of each modality;
\item A mutual-attentional feature fusion block is proposed to learn a conjoint feature representation for classification;
\item The proposed method achieves the state-of-the-art results on three standard RGB-D egocentric datasets.
\end{itemize}

\section{Related Work}
\textbf{Third Person RGB-D Action Recognition} \quad RGB-D based action recognition has attracted much attention and many works have been reported to exploit the complementary nature of RGB and depth modalities. Kong et al. \cite{kong2015bilinear} propose to project features from different modalities into shared space and learn RGB-Depth features for recognition. Wang et al. \cite{wangcv} take the scene flow as input and propose a new representation called Scene Flow to Action Map (SFAM) for action recognition. Instead of treating RGB and depth as separate channels, Wang et al. \cite{wangAAAI} propose to train a single CNN (called c-ConvNet) for RGB-D action recognition and a ranking loss is utilized to enhances the discriminative power of the learned features. Liu et al.~\cite{liujun} propose a multimodal feature fusion strategy to exploit geometric and visual features within the designed spatio-temporal LSTM unit for skeleton-based action recognition. Shahroudy et al.~\cite{2017deep} adopt a deep auto-encoder based nonlinear common component analysis network to discover the shared and informative components of input RGB+D signals. For more methods, readers are referred to the survey paper \cite{survey}. The above mentioned methods are mainly based on the third-person datasets, and the first-person action recognition has generated renewed interest due to the development of wearable cameras.

\par \textbf{First Person Action Recognition} \quad The early methods utilize semantic cues (object detection, hand pose and gaze information) to assist the egocentric action recognition. For example, a hierarchical model is presented by \cite{fathi} to exploit a joint representation of objects, hands and actions. Li et al. \cite{li2015delving} design a series egocentric cues for action recognition containing hand pose, head movement and gaze direction. The advance of the deep learning has led to the development of methods based on CNNs and RNNs. Several methods adopt two-stream structure \cite{two-stream} as the basic configuration and modify it to fit different purpose. Ma et al. \cite{ma2016going} redesign the appearance stream for hand segmentation and object localization. Singh et al. \cite{singh} propose a compact two-stream network which uses semantic cues. For temporal encoding, LSTM and ConvLSTM are employed in \cite{lsta,attention}. However these methods are all based on single RGB modality, and there are few works on RGB-D egocentric action recognition. Tang et al. \cite{THU-READ} propose a multi-stream network to incorporate features from RGB, depth and optical flow using Cauchy estimator and orthogonality constraint. Garcia-Hernando et al. \cite{garcia2018first} release a RGB-D egocentric dataset with hand pose annotation, however they do not propose any method based on RGB and depth. This paper focuses on first person action recognition using RGB and depth modalities.

\par \textbf{Transformer} \quad Transformer \cite{transformer}, a fully-attentional architecture, has achieved state-of-the-art results than RNN or LSTM based methods for sequence modeling problem e.g. machine translation and language modeling. Apart from NLP tasks, transformer has also been employed in some computer vision tasks such as image generation \cite{imagetransformer} and human action localization \cite{girdhar2019video}. Inspired by there works, we use transformer for intra-modality temporal modeling and cross-modality feature fusion.

\begin{figure*}
\begin{center}
\includegraphics[width=0.80\linewidth, height=80mm]{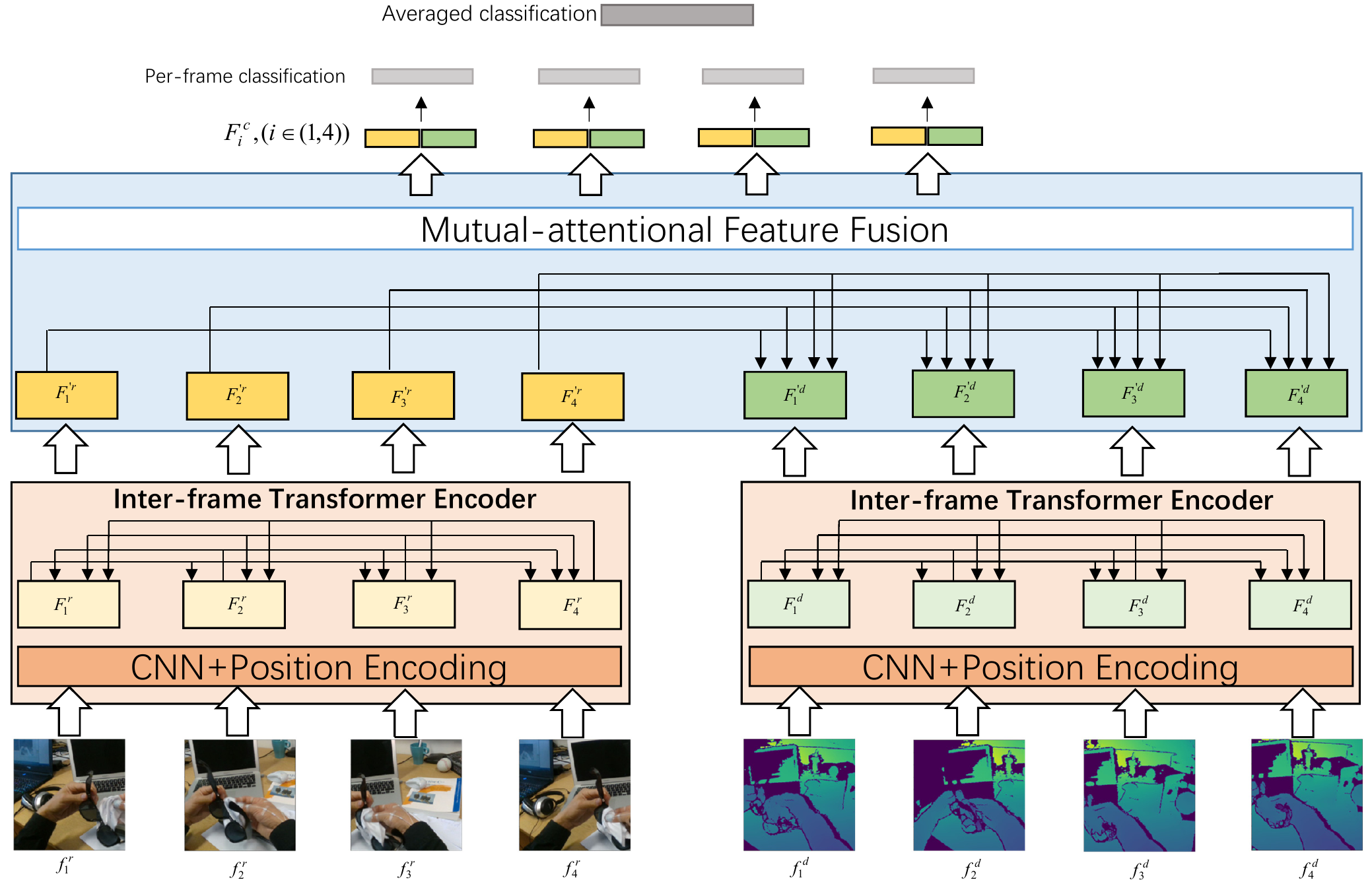}
\end{center}
   \caption{The illustration of the proposed framework. It takes four RGB frames and the corresponding depth maps as input, which are processed by two encoders respectively. Features from each modality are interacted and incorporated through the mutual-attentional block to produce the cross-modal or joint representation. The final classification is the average of each frames which produced by the joint representation.}
\label{Fig:1}
\end{figure*}

\section{Proposed Method}
In this section, we first give an overview of the proposed framework. Then both inter-frame transformer encoder and mutual-attentional feature fusion block will be described in detail.

\subsection{Overview}
The proposed method is developed for egocentric action recognition from heterogeneous RGB and depth modalities. As shown in Fig. \ref{Fig:1}, the proposed method contains two parts, two transformer encoders and a mutual-attentional fusion block. The network takes aligned RGB frames and depth maps as input, which are first converted into two sequences of feature embeddings. Then both sequence features are fed to the transformer encoders to model the temporal structure respectively. Features obtained from the encoders interact through the cross-modality block and then fused to produce the cross-modality representation. The conjoint features are processed through the linear layer to get per-frame classification and then averaged over the frames of an action clip as the final recognition result.

\subsection{Inter-frame Transformer Encoder}
\begin{figure}[t]
\begin{center}
\includegraphics[width=0.90\linewidth, height=38mm]{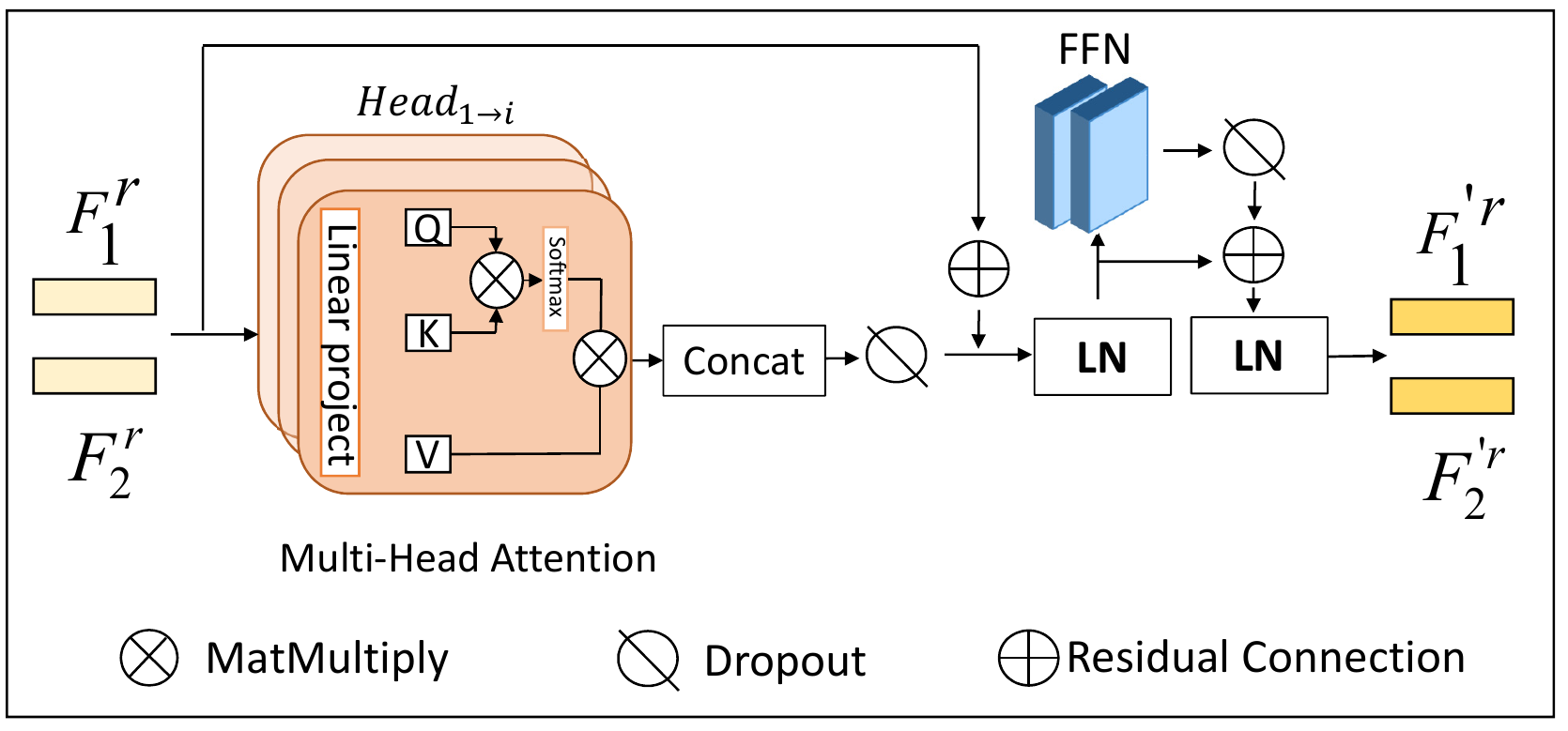}
\end{center}
   \caption{The architecture of the inter-frame transformer encoder. For simplicity, the number of feature embeddings are set to 2. Multi-head attention is first applied on the feature embeddings. Then the output of each head are concatenated and pass through series operations contains residual connection, drop and Layer Normalization (LN). FFN denotes Feed-Forward Network. Note that the calculation in the process is matrix operation which packing the embeddings.}
\label{Fig:2}
\end{figure}

As shown in Fig. \ref{Fig:1}, the two transformer encoders process both RGB and depth data respectively which form a two-stream structure. Since both streams are composed of the same network configuration (not weight-shared), here we just describe the RGB stream in detail. Given a sequence with $k$ RGB frames sampled from an action clip $\left\{f^r_1, f^r_2, ..., f^r_k \right\}$, we conduct average pooling on the feature maps of each frames to produce the feature embeddings $\left\{F^r_1, F^r_2, ..., F^r_k \right\}$ with size $d_{model}=512$. In order to encode the position information of each frame in the sequence, we utilize the position encoding proposed by Vaswani et al. \cite{transformer}, which adopting sine and cosine functions of different frequencies:
\begin{gather}
{PE_{(pos,2i)}} = sin(pos/10000^{2i/d_{model}})  \\
{PE_{(pos,2i+1)}} = cos(pos/10000^{2i/d_{model}}) 
\end{gather}
where $pos$ is the position and $i$ is the dimension. This function is chosen for the hypothesis that the model can easily learn to attend by relative positions, since for any fixed offset $k$, $PE_{pos+k}$ can be represented as a linear function of $PE_{pos}$. The positional encodings have the same dimension $d_{model}$ as the embeddings, so that the two can be summed. The remaining architecture essentially follows the standard Transformer which can be seen in Fig. \ref{Fig:2}. After obtaining the feature embeddings, multi-head attention is applied to them. Specifically, features are first mapped to a series vectors query $(Q)$ and key $(K)$ of dimension $d_k$, and value $(V)$ of dimension $d_v$ using different learned linear projection. Then the dot-product of the $Q$ vector and $K$ vector are calculated through softmax function to get the attention weight, and have a weighted sum of $V$. For each head, the process can be presented by

\begin{equation}
{\rm Head}_i = {\rm Softmax}(\frac{QK^T}{\sqrt{d_k}})V
\end{equation}
The concatenation of each head's output is followed by a group of operations containing dropout, residual connection and LayerNorm (LN) \cite{layernorm}.
\begin{gather}
{f'^r} = {\rm LN}(F^r+{\rm Dropout}({\rm Concat}({\rm Head}_i)))\\ 
{F'^r} = {\rm LN}({f'^r}+{\rm Dropout}({\rm FFN}({f'^r})))
\end{gather}

where ${F^r}$ is the matrices packing of input feature embeddings $\left\{F^r_1, F^r_2, ..., F^r_k \right\}$, $f'^r$ is the intermediate feature during process and ${F'^r}$ denotes the features after transformer encoder. Feed-Forward Networks (FFN) is composed of two convolution layers with kernel size being $1$. Notice that the whole process is based on the matrices calculation.

\par
\begin{figure}[!t]
\centering
\subfloat[Cropping same region of all frames.] {\label{Fig3a}\includegraphics[width = 0.99\columnwidth, height = 18mm]{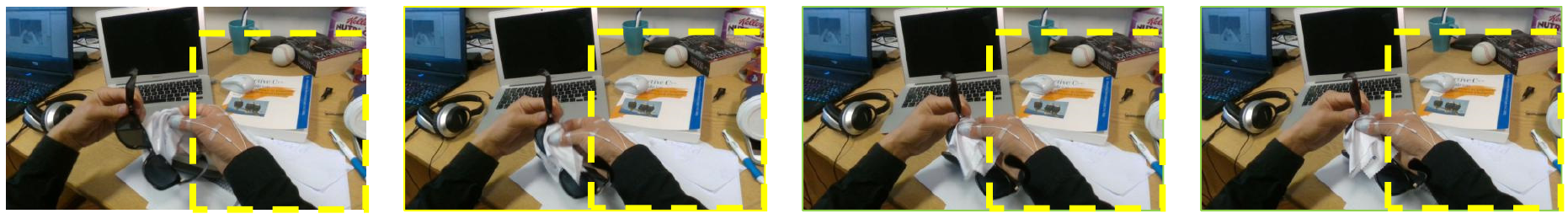}}\\
\subfloat[Cropping regions randomly of each sampled frames.] {\label{Fig3b}\includegraphics[width = 0.99\columnwidth, height = 18mm]{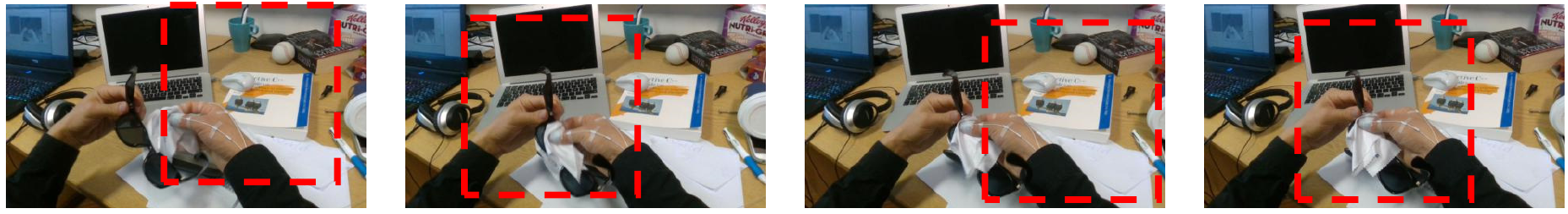}}\
\caption{Different data cropping manner, (a) yellow boxes indicates that regions are cropped at the same location as used in other methods. (b) red boxes indicates that regions are cropped randomly as adoped in our method. }
\label{Fig3}
\end{figure}

 Instead of using recurrent units, the inter-frame dependencies is modeled using self-attention mechanism. In order to enhance the spatial correlation, we adopt a simple yet effective image cropping operation. As we know, most RGB-based action recognition methods will apply the data augmentation (resize, crop, flip, etc.) randomly on the input images. For image cropping, 
as shown in Fig. \ref{Fig3}, the RGB images are sampled as a certain interval from the action clip (4 frames for simplify). Fig. \ref{Fig3a} shows the normal crop manner used in most methods \cite{lsta,attention}, where the yellow boxes indicate the cropped region of the original image, and the region are same for all the images. Fig. \ref{Fig3b} indicates the random cropping used in our method. For every sampled RGB image, we extract the region randomly. The red boxes show that the cropped regions can be located in image everywhere. In short, normal cropping only takes a fixed region of action videos into consideration during one training iteration, while our random cropping vice the verse. The random cropping has following advantages: 1) the egocentric action clips always short and have small range of motion, resulting plenty of inter-frames context redundancy. Comparing to crop same regions for all images, our cropping manner can augment the randomness of input data. 2) since the transformer is applied to model the inter-frame relationship, the repeated regions context will result in the inefficiency of attention calculation. Different cropped regions can effectively raise efficiencies and enhance the inter-frame spatial correlation. Results in Table \ref{table:5} shows the effectiveness of the proposed simple data operation.

\subsection{Mutual-attentional Feature Fusion}
\begin{figure}[t]
\begin{center}
\includegraphics[width=0.85\linewidth, height=45mm]{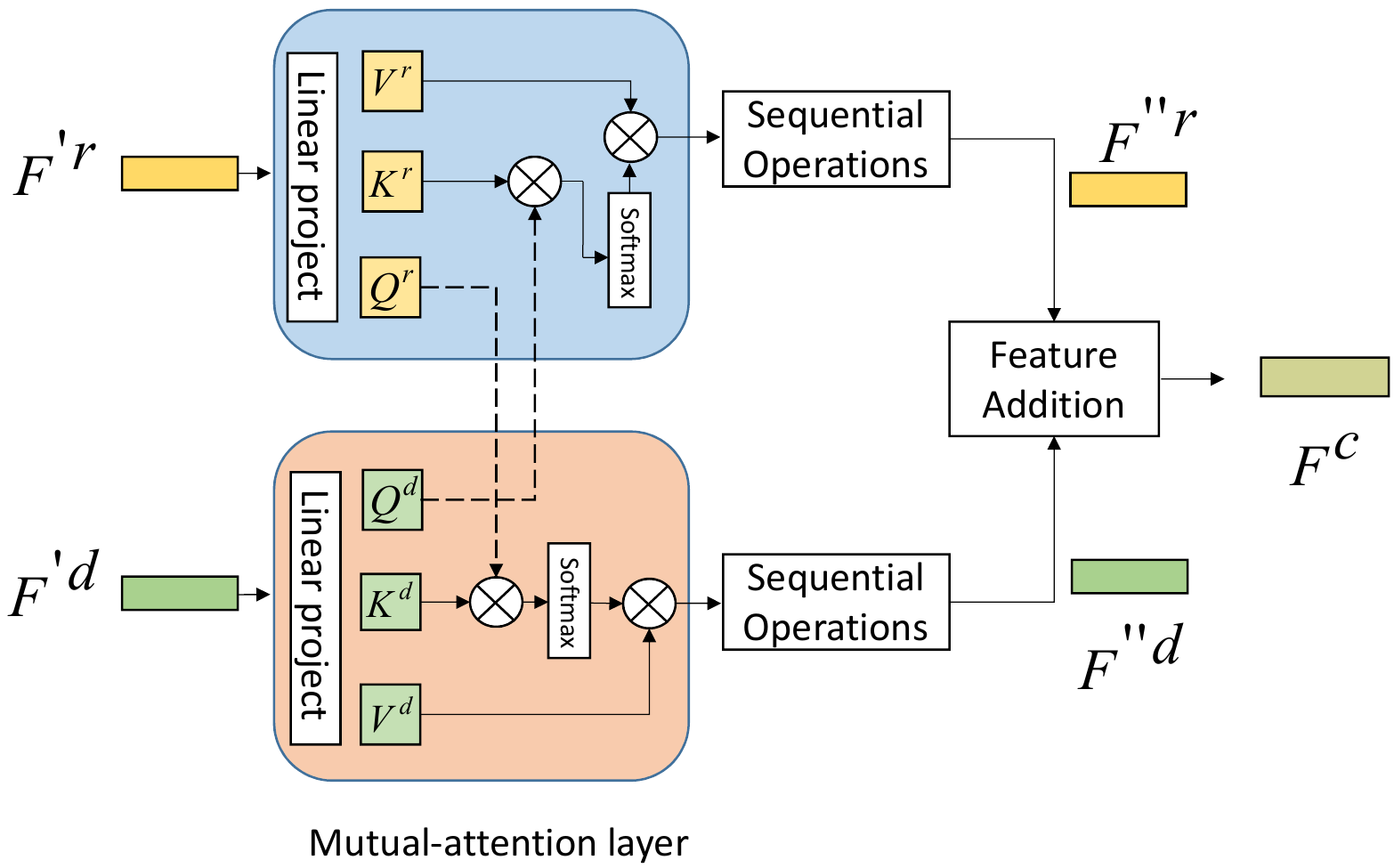}
\end{center}
   \caption{Illustration of the proposed mutual-attentional block. It consists of a mutual-attention layer and feature fusion operation. Feature embeddings from RGB and depth modalities are fed to the attention layer to exchange the information. Then the features are passing through the sequential operations similar to transformer encoder and then fused together to get the joint representation}
\label{Fig4}
\end{figure}

Due to the feature variations in different modalities, it's essential to learn a joint representation of the RGB and depth modalities. We propose a cross-modality block to interact features from both modalities using mutual-attentional mechanism. As shown in Fig.~\ref{Fig4}, the proposed module contains two parts, mutual-attention layer and feature fusion operation. The intermediate feature embeddings of both modalities from inter-frames encoder can be represented as $\left\{F'^r_1, F'^r_2, ..., F'^r_k \right\}$, $\left\{F'^d_1, F'^d_2, ..., F'^d_k \right\}$, where $r$ and $d$ represent RGB and depth. $Q^r$ ($Q^d$), $K^r$ ($K^d$) and $V^r$ ($V^d$) matrices are computed following the standard transformer. Then the mutual-attention is applied to retrieve the information from context vectors (key $Q^d$ and value $V^d$) of depth stream related to query vector $Q^r$ of RGB stream and vice the verse. Specifically, it calculates the RGB feature attention in depth modality and depth feature attention in RGB modality, and produces corresponding cross-modality features $\left\{F''^r_1, F''^r_2, ..., F''^r_k \right\}$ and $\left\{F''^d_1, F''^d_2, ..., F''^d_k \right\}$ respectively. Then the dropout, residual connection and LayerNorm operations are also employed sequentially. After mutual-attention layer, features of each frame from both modalities are fused via simply feature addition operation and used for per-frame classification. The final classification are the average of per-frame results. The whole process can be presented as follows

\begin{gather}
\left\{F''^r_i, F''^d_i \right\}  = {\rm MutAtten}(F'^r_i, F'^d_i), i \in (1, k)\\
F^c_i = F''^r_i + F''^d_i
\end{gather}

The mutual-attention mechanism builds the interaction among different modalities, and features incorporated after such layers can benefit from the narrow modality discrepancy than fusing the features extracted from modalities directly. Although such a co-attentional mechanism has been utilized in some vision-and-language tasks \cite{vilbert}, e.g. visual question answer (VQA) and visual commonsense reasoning (VCR). However, the co-attentional used in our method has two 
differences, 1) the two modalities data RGB and depth are restricted aligned, RGB frames and depth maps are one-to-one correspondence. For vision-and-language tasks, words and visual inputs often suffer from mismatch issue, which affect the attention computation among inputs. 2) the modality gap between visual feature and  word embedding are much complexity. While both RGB and depth are image-level visual feature, which make them interacted  with each other through mutual-attention layer more effectively and straightforward.

\begin{table*}[t]
\centering
\setlength{\tabcolsep}{0.5mm}{
\begin{tabular}{l|c|c|c}
\toprule
Methods                            &Modality                    & THU-READ (\%) & WCVS(\%) \\
\midrule
HOG \cite{HOG}                     & Depth                      & 45.83  &50.61 \\
HOF \cite{HOF}                     & Depth                      & 43.96  &41.25 \\
Depth Stream \cite{stream}         & Depth                      & 34.06  &58.47 \\
TSN \cite{TSN}                     & Depth                      & 65.00  &59.32 \\
\midrule
HOG \cite{HOG}                     & RGB                        & 39.93  &52.14   \\
HOF \cite{HOF}                     & RGB                        & 46.27  &48.50 \\
Appearance Stream \cite{stream}    & RGB                        & 41.90  &60.36 \\
TSN \cite{TSN}                     & RGB                        & 73.85  &66.02 \\
\midrule
TSN \cite{TSN}                     & RGB + Flow                 & 78.23  &67.05 \\
TSN \cite{TSN}                     & RGB + Flow + Depth         & 81.67  &70.09 \\
MDNN \cite{THU-READ}               & RGB + Flow + Depth + Hand  & 62.92  &67.04 \\
\midrule
Trear (Ours)                               & Depth                      &76.04   &63.72 \\
Trear (Ours)                                & RGB                        &80.42   &68.27 \\
Trear (Ours)                                & RGB+Depth                  &{\bfseries 84.90}  &{\bfseries 71.49}\\
\bottomrule
\end{tabular}}
\caption{Results obtained by the proposed ``Trear'' and comparison with the state-of-the-art methods on THU-READ and WCVS datasets. The results are the average of the 4 splits and 5 subjects respectively.}
\label{table:1}
\end{table*}

\section{Experiments}
The proposed method is extensively evaluated on three standard RGB-D action recognition benchmark datasets, the large THU-READ \cite{THU-READ} and FPHA \cite{garcia2018first} datasets and the small WCVS \cite{WCVS} dataset. Ablation studies and attention maps visualization are also reported to demonstrate the effectiveness of the proposed method.

\subsection{Implementation Detail and Training \label{Subsection:Implementation Detail}}
The proposed framework consists of two parallel streams corresponding to the RGB and depth modalities. They interact through the fusion block. Both inter-frame attention encoders share the same structure, and ResNet-34 pre-trained on the ImageNet dataset is adopted as the feature encoder. In order to reduce the computation cost, the number of encoder is set to 1. The number of the heads for attention calculation in both inter-frame block and mutual-attention block is set to 8. Notice that, we found that the number of heads for mutual-attention set to 2 can perform slightly better than 8 heads on THU-READ dataset, and we take the better results for THU-READ dataset comparison.
\par The experiments are conducted on the Pytorch framework with a single TitanX GPU. The networks are trained for 50 epochs with batch-size of 4 on all the three datasets. The initial learning rate is set to 0.0001 and the learning rate is decayed by a factor of 0.1 after 30 epochs. Adam optimizer is used to train all networks. For the input data, we select 32 frames from each action clip, uniformly sampled in time. Images are first resized to 256, and then randomly cropped to $224\times224$ for training. The center crop is used for testing. The depth data are first normalized to (0-1) and then copied into a 3-channel input so that the depth stream can directly utilize the pre-trained weight of ResNet-34.

\subsection{Datasets}
\par \textbf{THU-READ} \quad The THU-READ \cite{THU-READ} dataset is current the largest RGB-D egocentric dataset which consists of 40 different actions performed by 8 subjects. The RGB and depth data are collected by Primesense Carmine camera, which is a RGB-D sensor released by Primesense. It contains 1920 videos with each subject repeating each action for 3 times. We adopt the released leave-one-split-out cross validation protocol, which divides the 8 subjects into 4 groups and uses 3 splits for training and the rest for testing.

\par \textbf{FPHA} \quad The FPHA (First-Person Hand Action) \cite{garcia2018first} dataset collected with Intel RealSense SR300 RGB-D camera on the subject's shoulder. It contains 1175 sequences belonging to 45 action categories performed by 6 subjects. The dataset also has accurate hand pose annotation. It is separated into 1:1 setting for training and validation at video level with 600 sequences and 575 sequences respectively.

\par \textbf{WCVS} \quad Wearable Computer Vision Systems (WCVS) \cite{WCVS} dataset which is captured by RGB-D camera mounted on a helmet that contains three levels of action recognition. The Level 1 consists of two action categories, manipulation and non-manipulation. Level 2 subdivides the two action into 4 and 6 classifications respectively. Although Level 3 contains fine-grained actions, the recording frequency is too low to train a classifier. Following the \cite{WCVS,THU-READ}, we adopt Level 2 with 4 action classes to evaluate our method. The dataset is performed by 4 subjects in 2 scenarios. The large intra-class variations pose a great challenge to recognition. Cross-subject evaluation metrics is adopted in this paper.

\subsection{Results and Comparison with the State-of-the-art}

\begin{table}[!t]
\centering
\setlength{\tabcolsep}{0.01mm}{
\begin{tabular}{lcc}
\toprule
Methods                                              &Modality   & Accuracy \\
\midrule
Two stream-color \cite{two-stream-color}             & RGB   & 61.56    \\
H+O \cite{tekin2019h+}                               & RGB   & 82.43    \\
\midrule
${\rm HOG^2}$-depth \cite{ohn2014hand}               & Depth  & 59.83       \\
HON4D \cite{hon4d}                                   & Depth  & 70.61 \\
\midrule
2-layer LSTM \cite{garcia2018first}                  & Pose  &80.14\\
Gram Matrix \cite{GM}                                & Pose  &85.39\\
\midrule
Two stream  \cite{two-stream}                  & RGB + Flow  & 75.30 \\
${\rm HOG^2}$-depth+pose \cite{ohn2014hand}          & Depth + Pose  & 66.78 \\
\midrule
Trear (Ours)                                                  & Depth  &92.17\\
Trear (Ours)                                                  & RGB    &94.96\\
Trear (Ours)                                                  & RGB+Depth  &{\bfseries 97.04}\\

\bottomrule
\end{tabular}}
\caption{Results obtained by ``Trear'' and comparisons with the state-of-the-art methods on the FPHA dataset . Pose represents the hand pose modality.}
\label{table:2}
\end{table}

As shown in Table~\ref{table:1}, the compared methods mainly contain hand-crafted feature based methods: HOG \cite{HOG} and HOF \cite{HOF}, and deep learning-based methods: TSN \cite{TSN} and MDNN \cite{THU-READ}. In the cases of single modality, RGB-based methods perform better than depth-based methods because of the vital texture features that RGB modality carries. Benefited from the transformer, our method can explicitly model the intra-modality temporal structure and outperform others on the THU-READ and WCVS datasets. In the cases of multi-modality, TSN exploits optical flow modality to process the motion information and treats depth and RGB modality as separate channels for late score fusion. MDNN employs a multi-stream network and deploys Cauchy estimator and orthogonality constraint to assist egocentric action recognition. Our method achieves the state-of-the-art results, indicating that the learned conjoint cross-modal representation produced by mutual-attention block can effectively exploit the complementary nature of both modalities.

\begin{table}[t]
\centering
\begin{tabular}{lccc}
\toprule
Methods            &THU-READ  &FPHA   &WCVS  \\
\midrule
ResNet-34          &79.60     &89.16  &64.58 \\
ResNet-34+Encoder  &84.58     &94.96  &68.27 \\
\midrule
\end{tabular}
\caption{Ablation study for the Inter-frame Transformer Encoder on THU-READ (CS4), FPHA and WCVS datasets.}
\label{table:3}
\end{table}

\par Since FPHA can be adopted as hand pose estimation benchmark, thus hand pose annotations are given as a known modality and can be used for action recognition as well. It can be seen from Table \ref{table:2}, methods based on hand pose outperform most of those based on RGB and/or depth. Since the egocentric video mostly contains the hands and interacted objects, the hand pose feature contributes significantly to the recognition performance. Tekin et al. based on this characteristic, \cite{tekin2019h+} develop a unified framework that can estimate 3D hand, object poses and action category from RGB data. Two-stream \cite{two-stream} utilizes the optical flow to exploit short-term motion information and \cite{garcia2018first} introduces the temporal information vis recurrent units (LSTM). Benefit from the proposed intra-frame encoder, our method can process the input frames parallel and build the context correlation of the action clip without using flow and recurrent unit. In short, our method (single modality or both RGB-D) outperforms all other methods by a large margin, demonstrating the effectiveness of both transformer encoder and mutual-attention block in fine-grained egocentric action recognition.

\begin{table}
\centering
\begin{tabular}{lccc}
\toprule
Methods            &THU-REA  & FPHA  & WCVS\\
\midrule
\multicolumn{4}{c}{Single Modality}\\
\midrule
Depth               &77.50  &  92.17  &  63.72 \\
RGB                 &84.58  &  94.96  &  68.27 \\
\midrule
\multicolumn{4}{c}{RGB+D Feature Fusion}\\
\midrule
Concatenation       &86.25  & 94.43 & 70.16  \\
Multiplication      &86.67  & 96.00 & 69.60  \\
Addition            &86.67  & 94.09 & 69.59  \\
\midrule
\multicolumn{4}{c}{RGB+D Mutual-attentional Feature Fusion}\\
\midrule
Concatenation       &86.67  & 95.30 & 70.23  \\
Multiplication      &85.83  & {\bfseries97.04} & 69.58   \\
Addition            &{\bfseries 88.33}  & 96.34  & {\bfseries 71.50} \\
\bottomrule
\end{tabular}
\caption{Ablation study of the proposed mutual-attention fusion block with different fusion manners on THU-READ (CS4), FPHA and WCVS dataset.}
\label{table:4}
\end{table}

\subsection{Ablation Studies}
In order to verify the effectiveness of the proposed inter-frame Transformer encoder and fusion block, ablation studies are conducted on the THU-READ, FPHA and WCVS datasets. Since the inter-frame Transformer encoder in our framework is composed of CNN and Transformer encoder, we conduct an the ablation study for the ResNet-34 and the encoder, as shown in the Table \ref{table:3}. The results shows that the Transformer encoder can effectively model the temporal structure and can improve the performance significantly. As shown in Table \ref{table:4}, RGB modality contributes more significantly to the recognition than depth modality because of the needed textual information. Directly fusing features from both modalities for recognition even produces slightly worse results than using RGB alone, especially on the FPHA dataset. This is probably due to the neglect of the modality discrepancy. The proposed mutual-attentional block can effectively mitigate such an issue, in which the features from different modalities can exchange the information through the mutual-attention layer to reduce the feature variations. Then the cross-modal features are fused to produce the conjoint feature representation. The results also show that the addition fusion performs better than concatenation and multiplication fusion on THU-READ and WCVS and a slightly worse in FPHA. In addition, Table \ref{table:5} shows the results of different image cropping methods, indicating that random cropping used in the proposed method improves the performance significantly.

\begin{table}[t]
\centering
\begin{tabular}{lcc}
\toprule
Methods            &Random Crop         & Crop same region \\
\midrule
\multicolumn{3}{c}{THU-READ}\\
\midrule
Depth               &77.50              &  73.75  \\
RGB                 &84.85              &  82.08  \\
RGB+Depth           &88.33              &  86.67  \\
\midrule
\multicolumn{3}{c}{FPHA}\\
\midrule
Depth               &92.17              &  88.00 \\
RGB                 &94.96              &  91.65  \\
RGB+Depth           &97.04              &  95.65  \\
\midrule
\end{tabular}
\caption{Ablation study for the data crop manner on THU-READ (CS4) and FPHA datasets.}
\label{table:5}
\end{table}

\begin{figure}[!t]
\centering
\subfloat[Attention map in RGB stream transformer encoder.]{\label{Fig5a}\includegraphics[width = 0.70\columnwidth, height = 33mm]{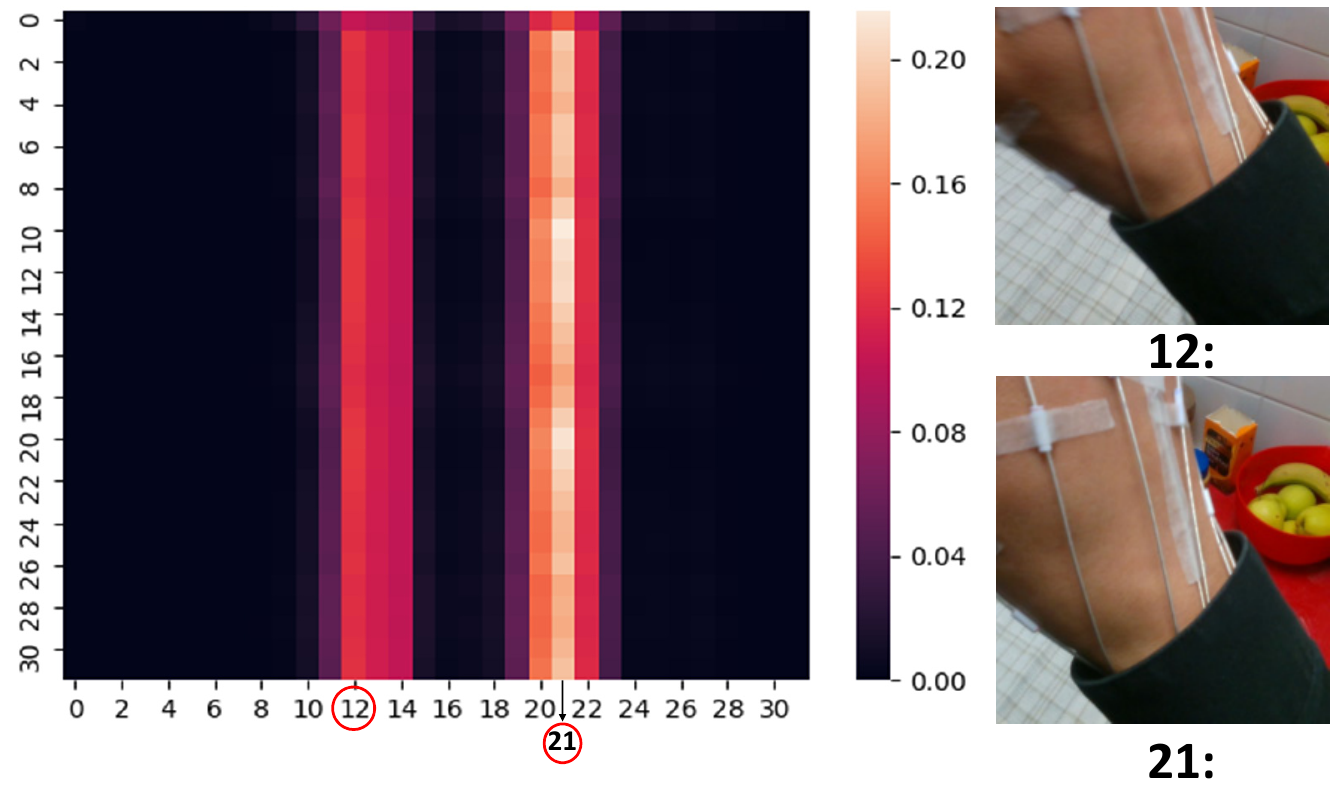}}\\
\subfloat[Attention map in mutual-attentional layer.]{\label{Fig5b}\includegraphics[width = 0.90\columnwidth, height = 32mm]{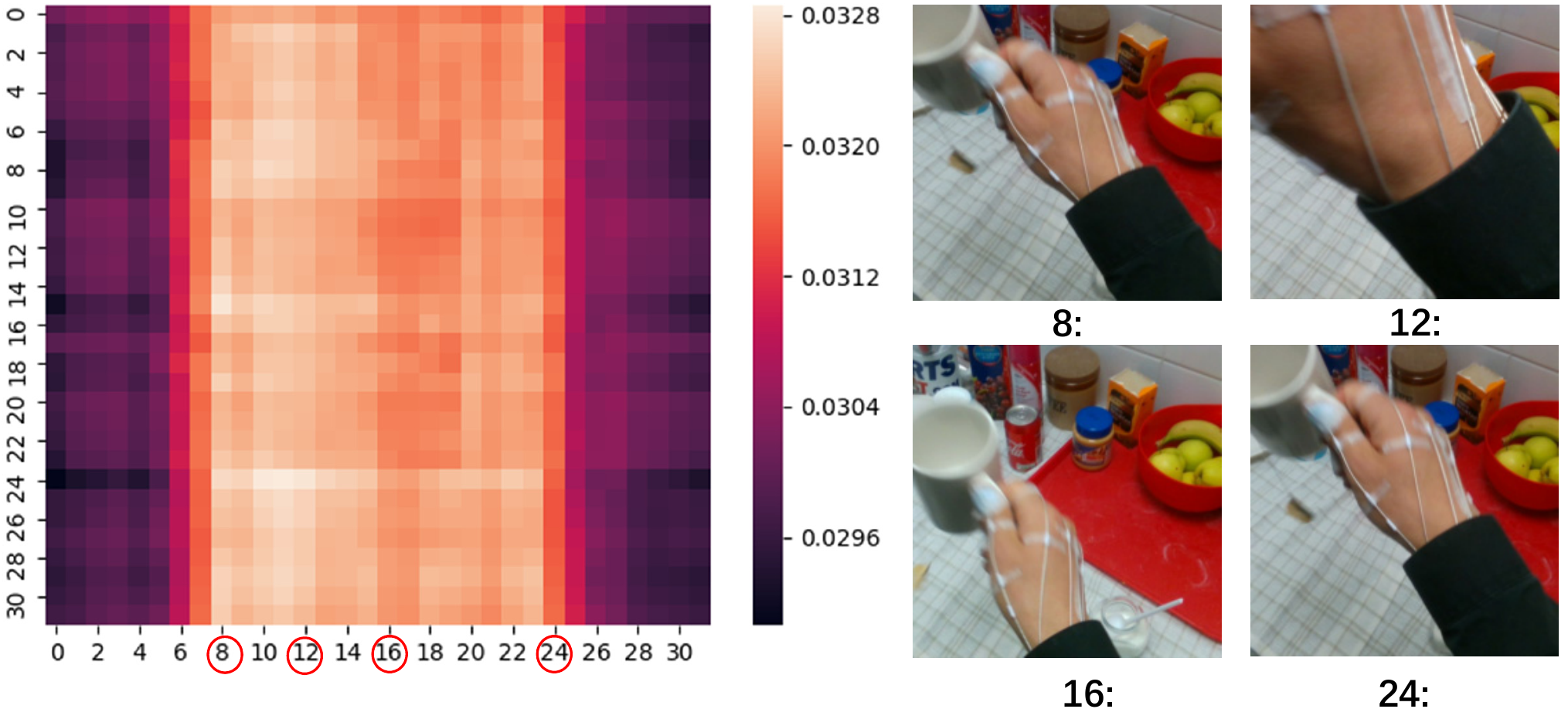}}\
\caption{Attention maps in both inter-frames attention and mutual-attentional layers. The vertical axis denotes the query vectors and the horizontal axis represents the context vectors. The action indicates ''drink mug'' in FPHA dataset.}
\label{Fig5}
\end{figure}

\subsection{Attention Map Visualization}
As shown in Fig. \ref{Fig5}, the attention maps in inter-frame transformer encoder and mutual-attentional layer are visualized respectively. The action is the ''drink mug'' with 32 sampled frames, which conducts "drink" process twice, $0-14$ frames is the first ''drink'' and $15-26$ is the second ''drink''. Fig. \ref{Fig5a} represents the inter-frame temporal attention weight in the RGB stream transformer encoder. It can be seen that the encoder can accurately capture the "drinking" moments and most of frames are correlated to the two moments. However, the correlation to the actions ''Pick up mug'' and ''put down mug'' are weak which mainly because of the appearance change are small in these actions. Fig. \ref{Fig5b} shows the attention map in mutual-attentional layer. From the figure, we can see that the proposed co-attention mechanism can exploit the complementary characteristics of both modality and model the complete action conduction process, contains "Pick up mug $-$ drinking $-$ put down mug" twice.

\section{Conclusion}
In this paper, we present a novel framework for egocentric RGB-D action recognition. It consists of two modules, inter-frame transformer encoder and the mutual-attentional cross-modality feature fusion block. The temporal information is encoded in each modality through the self-attention mechanism. Features from different modalities can exchange information via the mutual-attention layer and fused to become the conjoint cross-modal representation. Experimental results on three RGB-D egocentric datasets demonstrates the effectiveness of the proposed method.

\section*{Acknowledgment}
This work was supported in part by the National Natural Science Foundation of China (Grant numbers: 61906173, 61822701).

\end{document}